\documentclass{article}
\usepackage{amssymb}
\usepackage{lipsum}

\usepackage{multirow}
\usepackage{graphicx}
\usepackage{subcaption}
\renewcommand{\vec}[1]{\mathbf{#1}}

\pdfoutput=1 %PER ARXIV
\usepackage{amssymb}
\usepackage{arydshln}
\usepackage[normalem]{ulem}
\usepackage{subcaption}
\usepackage{booktabs}

\usepackage{tikz}
\usetikzlibrary{positioning, shapes, shadows, arrows}
\usetikzlibrary{decorations.pathreplacing}
\tikzstyle{abstract}=[circle, draw=black, fill=white]
\tikzstyle{labelnode}=[circle, draw=white,opacity=.2,text opacity=1]
\tikzstyle{invisiblenode}=[circle,dashed, inner sep=1pt,circle split,line width=1mm,minimum size=1.5cm]
\tikzstyle{line} = [draw, -latex']
\usepackage[utf8]{inputenc}

\title{}
\author{}
\date{}

\usepackage{graphicx}

\title{Towards a general framework for improving the performance of classifiers using XAI methods}
\author{Andrea Apicella, Salvatore Giugliano, Francesco Isgrò, Roberto Prevete\\
\textit{\small Department of Electrical Engineering and Information Technology, University of Naples Federico II}}
\date{}

\begin{document}

\maketitle

\begin{abstract}
Modern Artificial Intelligence (AI) systems, especially Deep Learning (DL) models, poses challenges in understanding their inner workings by AI researchers. eXplainable Artificial Intelligence (XAI) inspects internal mechanisms of AI models providing explanations about their decisions. While current XAI research predominantly concentrates on explaining AI systems, there is a growing interest in using XAI techniques to automatically improve the performance of AI systems themselves.
This paper proposes a general framework for automatically improving the performance of pre-trained DL classifiers using XAI methods, avoiding the computational overhead associated with retraining complex models from scratch.
In particular, we outline the possibility of two different learning strategies for implementing this architecture, which we will call \textit{auto-encoder}-based and \textit{encoder-decoder}-based, and discuss their key aspects.
%Our proposal is evaluated using three widely recognized DL models (EfficientNet-B2, MobileNet, and LeNet-5) along with three standard image datasets: CIFAR-10, CIFAR-100, and STL-10. The results show that IMPACTX consistently improves the performance of all pre-trained DL models across all evaluated datasets.

\end{abstract}

\section{Introduction}
Understanding the decision-making processes of modern Machine Learning (ML) and Deep Learning (DL) models remains a significant challenge. Explainable Artificial Intelligence (XAI) aims to address this challenge by providing possible explanations about inner workings of AI models, helping users in understanding the reasoning behind AI decisions. XAI techniques have been applied across various domains, including image analysis, natural language processing, and clinical decision support systems.

Despite the advancements in providing explanations for AI systems, there has been relatively less emphasis on adopting XAI to directly enhance the performance of ML models \cite{weber2022beyond}. Existing approaches often require human intervention and may not fully exploit the potential of XAI for improving model performance (e.g., Explanatory Interactive Learning, \cite{schramowski2020making}).

To address these limitations, we introduce a novel framework designed to automatically enhance the performance of pre-trained classifiers without extensive retraining. The proposed framework leverages XAI techniques to improve the decision-making capabilities of pre-existing models. In particular, by integrating XAI-derived explanations with the output of pre-trained classifiers, our framework want to improve model performance, furthermore without computationally intensive model retraining processes.

In this paper, we present a conceptual framework able to to bridge the gap between XAI and model performance enhancement.
\section{Related works}
\label{sec:related}
From a broader perspective, eXplainable Artificial Intelligence (XAI) methods can incorporate external knowledge from data to enhance Machine Learning (ML) systems, as shown in studies such as \cite{hu-etal-2016-harnessing,apicella2018integration}. By integrating Deep Neural Networks (DNNs) with human-derived domain knowledge, it becomes possible to lead data-driven learning approaches away from producing counter intuitive outcomes \cite{hu-etal-2016-harnessing}. Notably, in works like \cite{teso2019explanatory,schramowski2020making}, a mechanism called eXplanatory interactive Learning (XiL) is introduced, allowing for user interaction during the training phase to shape the model's outputs based on visual explanations. This interactive framework presents predictions along with explanations, enabling users to refine the model as needed during training. The efficacy of different types of user feedback within XiL on model performance and explanation accuracy is investigated in detail in \cite{hagos2022impact}.

However, integrating XAI methods into a ML pipeline to automatically enhance performance represents a relatively novel direction in current research. For example, \cite{NEURIPS2023_9c537882} introduces a theoretical framework analyzing how explanations, serving as a form of knowledge, can enhance model training. 
Differently, ``Right for the Right Reasons (RRR)'' \cite{ijcai2017p371} consisted in constraining the training loss function with proper regularization terms acting on the model gradients w.r.t. the inputs. 
The authors of \cite{apicella2023strategies} made a preliminary study on a family of XAI methods available in the literature to build features' relevance masks. Instead, in \cite{sun2021explanation} LRP explanations \cite{montavon2019layer}  led a ML model to focus on the important features during the training stage of a few-shot classification task. The effectiveness of SHapley Additive exPlanations (SHAP) \cite{NIPS2017_7062} values in feature selection strategies is investigated in \cite{marcilio2020explanations, gramegna2022shapley}. Conversely, in \cite{arslan2022towards} using SHAP values as input features for ML classification tasks is investigated. The authors reported that SHAP values can be adopted as input features to obtain model with improved performance respect to standard features provided in several publicly available datasets. \cite{fukui2019attention} introduced an attention mechanism in a CNN as a further neural network branch (Attention Branch Network, ABN) that focuses on a specific region in an image extending a visual explanation model. In \cite{mitsuhara2019embedding} the attention branch of ABNs are constrained to output the same attention maps of the ones edited by human experts.
In \cite{schiller2019relevance}, Deep Taylor Decomposition (DTD) relevance \cite{montavon2017explaining} is used to build a reliable classifier for detecting the presence of orca whales in hydrophone recordings. In particular, the DTD relevance is used as a binary mask to select the most relevant input features. In \cite{bento2021improving} LRP was adopted and manually examined to remove unnecessary information from the data.

Importantly, \cite{weber2022beyond} reports a survey of works using XAI approaches to enhance classification systems. In particular, \cite{weber2022beyond} categorizes current strategies for improving ML models with XAI into four main categories: i) augmenting the data, explanations are utilized to generate artificial samples that provide insights into undesirable behaviors (e.g., \cite{teso2019explanatory,schramowski2020making}); ii) augmenting the intermediate features, this involves utilizing feature-wise information provided by explanations to assess the importance of intermediate features (e.g.,\cite{anders2022finding,fukui2019attention,apicella2023strategies,apicella2023shap}); iii) augmenting the loss, additional regularization terms based on explanations are incorporated into the loss training function  (e.g.,\cite{ijcai2017p371,liu-avci-2019-incorporating}); iv) augmenting the model, these approaches entail modifying a model based on an estimation of the importance of its parameters (e.g., \cite{yeom2021pruning}).

Although this proposal is related to categories ii), iii) and iv), we point out that based on this categorisation, the predominant trend is to use XAI when training an ML model from scratch. In contrast, our proposal is tailored to improve an already trained ML model.

\section{Method Description}
\label{sec:method2}

In this work, we propose a strategy for integrating explanations into model predictions. Given a dataset comprising $n$ labeled instances $D = \{(\vec{x}^{(i)}, y^{(i)})\}_{i=1}^n$ and $m$ unlabeled instances $U =\{\vec{x}^{(j)}\}_{j=1}^m$, our objective is to predict the correct class of the elements of $U$ using a ML model $M$ trained on $D$.

Our proposed approach introduces a model $F$ aimed at identifying relevant input features for the true class $y^{(i)}$ to facilitate the classification task. Let $\vec{m}^{(j)} = M(\vec{x}^{(j)})$, where $\vec{m}^{(j)}$ represents the output given by $M$ for an input $\vec{x}^{(j)}\in S^E$, and $\hat{y}^{(j)}_M=\arg\max\big({softmax}\big(\vec{m}^{(j)}\big)\big)$ denotes the estimated label of the sample $\vec{x}^{(j)}$. Concurrently, we introduce an 'attribution encoding', denoted as $\vec{f}^{(j)} = F(\vec{x}^{(j)})$, designed to capture crucial information regarding the attribution of the input features $\vec{x}^{(j)}$ to the true class $y^{(j)}$. We hypothesize that $\vec{f}^{(j)}$, when combined with the output $\vec{m}^{(j)}$ generated by $M$, effectively aids in classifying the input $\vec{x}^{(j)}$ assembling them together through an additional simple classification model that has been properly trained.

\subsection{Training phase}
\label{sec:trainPhase}
A critical issue is how to train the $F$ module. Different learning strategies could be considered and carefully tested. In this paper we raise the possibility of two different strategies, which we will call \textit{auto-encoder} based and \textit{encoder-decoder} based. The first one can be outlined by the following four consecutive main steps: 1) Each input data $\vec{x}$ belonging to the training set is sent to the pre-trained classifier $M$ and a corresponding explanation $e_x$ of the classifier's behaviour is generated; 2) All the explanations $e_x$ are sent to an auto-encoder, thus generating an encoded explanation $z_x$; 3) the $F$ module is trained on the pair $(x, z_x)$; 4) at the end of this training phase, the simple classifier $C$ is trained using $z_x$ and the $M$ output. In contrast, the encoder-decoder based strategy can be described as follows: an encoder-decoder is trained directly using the pairs $(\vec{x},e_x)$ as a training set. During training, the hidden values $z_x$ of the encoder-decoder are sent as input to the simple classifier $C$ togheter with the $M$ output, thus training the classifier $C$ at the same time.  
\section{Conclusions}
\label{sec:conclusion}
In conclusion, it would be valuable to investigate the impact of the proposed approach on the automatic improvement of an already trained ML model. As discussed in Section \ref{sec:method2}, a first issue to be addressed is how to obtain, through a learning approach, the module $F$ that should provide important information about the input features to improve the classification. We have raised the possibility of two different types of learning schemes for this purpose: \textit{auto-encoder} and an \textit{encoder-decoder} based. A careful experimental analysis of these schemes should be carried out. 
A notable aspect of our approach is that we can directly obtain the explanation of the model responses without using an XAI method. This is achieved through $F$ and the $D$ decoder.

Moreover, it could be valuable to explore the impact of this approach by employing different XAI methods, architectures, and datasets than those used to train model $M$. 
For example, notice that in training the module $F$ one could assume that the original training set that was used to train the model $M$ is available. This hypothesis could be relaxed by assuming that we only have a part of this data set or another similar data set. Therefore, investigating the robustness of our approach with respect to a modification of the original dataset used to train the model $M$ is certainly a relevant aspect of this approach.

Finally, we aim to investigate the potential of our approach by gradually integrating the contribution of module $F$ during the learning phase while retraining the entire architecture alongside $M$ from scratch.

\bibliographystyle{plain}
\bibliography{bib}
\end{document}